\documentclass[nohyperref]{article}

\usepackage{microtype}
\usepackage{graphicx}
\usepackage{subfigure}
\usepackage{booktabs} % for professional tables

\usepackage{hyperref}

\usepackage[accepted]{icml2022}

\usepackage{amsmath}
\usepackage{amssymb}
\usepackage{mathtools}
\usepackage{amsthm}

\usepackage[capitalize,noabbrev]{cleveref}

\theoremstyle{plain}

\theoremstyle{definition}

\theoremstyle{remark}

\usepackage[textsize=tiny]{todonotes}

\icmltitlerunning{Submission and Formatting Instructions for ICML 2022}

\begin{document}

\twocolumn[
\icmltitle{Computational Co-Design for Variable Geometry Truss}



\icmlsetsymbol{equal}{*}

\begin{icmlauthorlist}
\icmlauthor{Jianzhe Gu}{cmu}
\icmlauthor{Lining Yao}{cmu}
\icmlcorrespondingauthor {Lining Yao}{liningy@andrew.cmu.edu}
\end{icmlauthorlist}

\icmlaffiliation{cmu}{Human-Computer Interaction Institute, Carnegie Mellon University, Pittsburgh, USA}

\icmlkeywords{Computational Design, Co-design, Robotics, Shape-changing Interface, Reinforcement Learning, Graph Neural Network}

\vskip 0.3in
]



\printAffiliationsAndNotice{}  % leave blank if no need to mention equal contribution

\begin{abstract}
Living creatures and machines interact with the world through their morphology and motions. Recent advances in creating bio-inspired morphing robots and machines have led to the study of variable geometry truss (VGT), structures that can approximate arbitrary geometries and has large degree of freedom to deform. However, they are limited to simple geometries and motions due to the excessively complex control system \cite{truss2, trussformer}. While a recent work PneuMesh \cite{gu2022pneumesh} solves this challenge with a novel VGT design that introduces a selective channel connection strategy, it imposes new challenge in identifying effective channel groupings and control methods. 

Building on top of the hardware concept presented in PneuMesh, we frame the challenge into a co-design problem and introduce a learning-based model to find a sub-optimal design. Specifically, given an initial truss structure provided by a human designer, we first adopt a genetic algorithm (GA) to optimize the channel grouping, and then couple GA with reinforcement learning (RL) for the control. The model is tailored to the PneuMesh system with customized initialization, mutation and selection functions, as well as the customized translation-invariant state vector for reinforcement learning. The result shows that our method enables a robotic table-based VGT to achieve various motions with a limited number of control inputs. The table is trained to move, lower its body or tilt its tabletop to accommodate multiple use cases such as benefiting kids and painters to use it in different shape states, allowing inclusive and adaptive design through morphing trusses. 

\end{abstract}

\section{Introduction}
\label{introduction}

Living creatures interact with the world through a large variety of transformation behaviors. Caterpillars repetitively shrink and twist the body to climb; pufferfish expand their bellies to intimidate enemies; and macrophages deform the cytomembrane to swim and hunt other cells. In the field of robotics, one robotic system that can create complex geometries and motions is Variable Geometry Truss (VGT) \cite{truss1, truss2, trussformer}. A VGT typically consists of an arbitrary number of beams and joints, where beams are connected through joints and form truss structures composed of tetrahedrons. Each beam is a linear actuator that can expand or contract. By morphing the body with actuators, VGTs can execute various motions, including rotation, twisting, linear and volumetric scaling.

\begin{figure}[ht]

\begin{center}
\centerline{\includegraphics[width=\columnwidth]{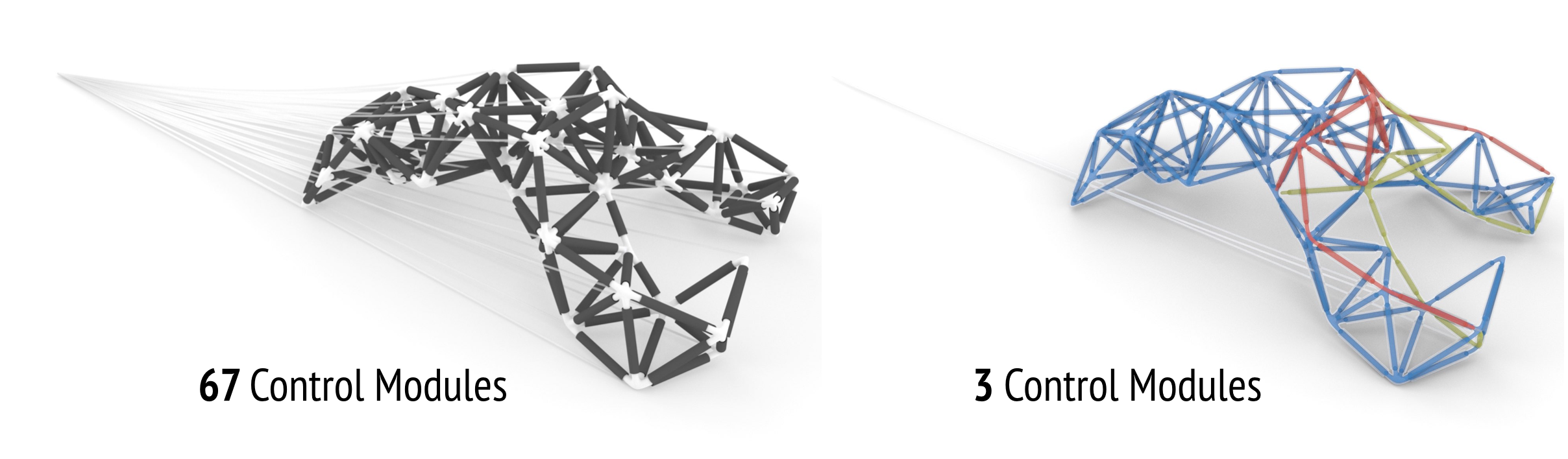}}
\caption{Left: A lobster VGT with individually controlled actuators. Right: A lobster VGT with channel grouping mechanism.}
\label{fig1}
\end{center}

\end{figure}

Despite its versatility and adaptability, a VGT suffers from the exponentially scaled complexity of the control system. As each beam is controlled independently, the number of control units (e.g. air tubings or linear motors) is cubically scaled with the complexity of the truss morphology. A recent VGT design, PneuMesh \cite{gu2022pneumesh}, simplifies the control system while achieving various complex motions. A PneuMesh VGT is a pneumatic driven truss where each beam is a syringe-like linear actuator with an air channel inside. Rather than individually controlling beams, PneuMesh selectively connects the air channels of beams through multi-way joints that direct air to separate beams. With this design, beams are integrated into different sub-networks, each sub-network are controlled synchronously by a single air valve. With PneuMesh, a complex VGT (up to more than 118 beams) can be controlled by a limited number of control modules (three to six control modules) but still create rich motions through the connections of beams. For example, a lobster-shaped VGT (Figure \ref{fig1}) with 67 beams can move forward and grab objects (Figure \ref{fig2} c) with only three control modules (Figure \ref{fig2}a).

Despite the simplification of the control system, identifying an optimal channel grouping and control strategy in a forward design or manual fashion is challenging. A truss with $n_e$ edges and $n_\gamma$ channels will have a solution space with a size up to $n_\gamma ^ {n_e}$. Meanwhile, a physically practical solution must keep the connectivity of each channel (i.e., beams belong to the same channel group need to have ensured physical connectivity for pneumatic actuation system). Lastly, the motion performance is very sensitive to the grouping strategy, where changing the channel of a single beam might lead to entirely different motions. Other than designing the channel assignment, for every channel design and target motion, a control strategy needs to be computed. As the number of beams and tasks increases, it is not practical to manually design both the channel assignment and the control. 

The aforementioned challenges lead to our proposed VGT co-design strategy. Given a predefined truss structure, we want to find a channel assignment and a set of control policies that enables the VGT to achieve multiple morphing tasks. Researchers have explore the problem of optimizing both the morphology and control of irregular-shaped robots. \cite{cubicrobot} use CPPN and genetic algorithms for a periodic control signal. \cite{zhao2020robogrammar} presented a graph heuristic search and reinforcement learning for co-designing a tree-structure robot.
While these work achieved great performance for their specific systems, they have different inherent topology from PneuMesh VGT which is a 3D graph with loops and specific constraints on the channel grouping.

In our work, we propose a hybrid model to solve the co-design problem of PneuMesh VGTs through genetic algorithms (GA) and reinforcement learning (RL). We adopted NSGA-II for the selection method of GA to improve the performance of multiple objectives without being biased on one of the objectives. Customized initialization and mutation functions are implemented to ensure the symmetry and the connectivity of channels. To evaluate the potential of a channel design, we train a control policy with a RL controller using PPO \cite{ppo} method, and simulate the motion of a VGT controlled by the trained policy. The simulation results are sent into customized objective functions to generate the ratings, or fitness. 

We demonstrate the effectiveness and efficiency of our method by designing a robotic truss table and applying our method on training the table towards four objective motion tasks, including lowering the height, tilting the table top, locomotion and turning around.

\section{Physical System}
In the previous work, PneuMesh \cite{gu2022pneumesh} introduces a VGT design that can minimize the number of control units while keeping a sufficient degree of control over the truss and achieving various motion tasks.

\begin{figure}[ht]
% \vskip 0.2in
\begin{center}
\centerline{\includegraphics[width=\columnwidth]{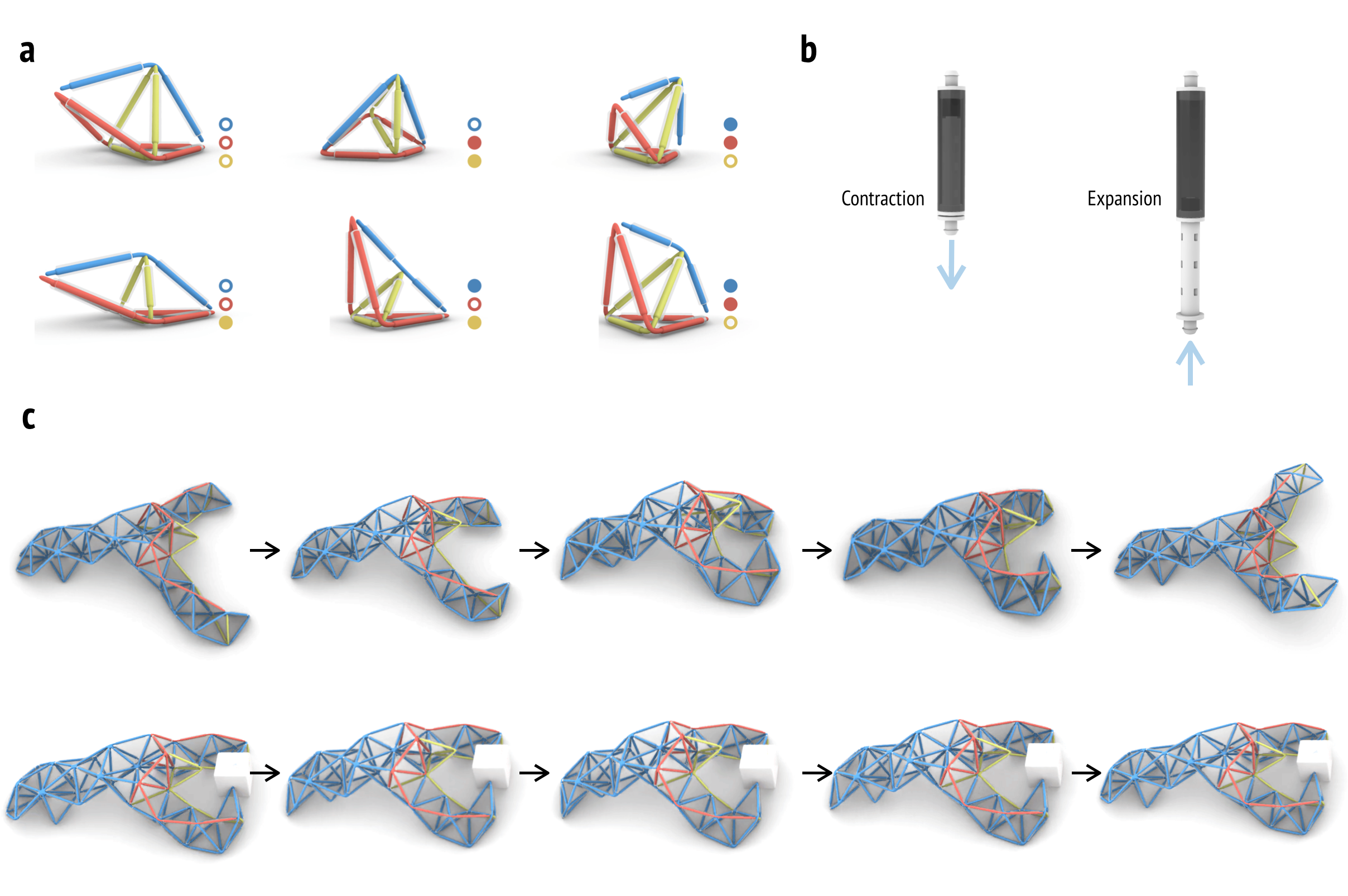}}
\caption{Physical system of channel grouped variable geometry truss.}
\label{fig2}
\end{center}
% \vskip -0.2in
\end{figure}

    Traditional VGTs are over-actuated systems where the number of beams is more than what is needed for given motion tasks. Instead of having independent actuators, we selectively connect adjacent actuators through their shared joints with a customized inner air channel design (Figure \ref{fig2}a,b). Interconnected actuators are synchronized and will be actuated or de-actuated at the same time corresponding to a single control input.
    
    In such a way,a system with $n_e$ edges and $n_c$ channels will have $n_c$ binary values as control inputs at every moment. Despite the decrease in the degree of freedom(DOF), each beam can still change length, and each channel is a sub-network that can span through the entire truss. These properties endow the truss with substantial freedom to deform and achieve various motions.

A trade-off exists between the reduction of DOF and simplification of the control system. The goal of this work is to find a balance between optimizing the channel assignment and control signals.

\section{Method}

\subsection{Definition and Problem Statement}

We define a truss structure as a graph $T = \{E, n_\gamma\}$ consisting of $n_v$ vertices and $n_e$ edges, where $E$ is the adjacency list of vertices and $n_\gamma$ is the predefined number of channels. A $T$ is associated with the channel assignment of its edges $C = [c_0, c_1, ... c_{n_e-1}]$, where $c_{i}$ is the integer index of the channel assigned to the $i\text{th}$ edge.
    
At every time step $t$, $T$ has a corresponding state $s_t = \{V_t, U_t, {V_t}', {U_t}', k_t\}$ , where $V_t$ and ${V_t}'$ are absolute and relative node positions, $U_t$ and ${U_t}'$ being absolute and relative node velocities. The actuation state of channels is a vector of $n_\gamma$ Boolean values, each indicating the on/off state of one channel. 

A simulator $M$ takes in $T$, $C$ and $s_t$, and outputs the next state $s_{t+1}$. $T$ can be assigned with multiple objective functions $O = [o_0, o_1, ... o_{n_o - 1}]$ that takes in the state sequence of an entire simulation $S = [s_0, s_1, s_2, ...]$ and outputs a rating vector $R = [r_0, r_1, ... r_{n_o - 1}]$, where each $o_i$ takes in $S$ and outputs $r_i$. A control policy $\pi$ takes in $T$, $C$, $s_t$, and outputs an action $a_t$, where $a_t$ is assigned to each channel as the new channel states. 

Given a $T$, its initial state $s_0$, and $O$, we want to find an optimal channel assignment $C$ and control policies $\Pi = [\pi_0, \pi_1, ... \pi_{N_o - 1}]$ that maximize the general performance on all objectives.

\begin{figure}[ht]
\begin{center}
\centerline{\includegraphics[width=\columnwidth]{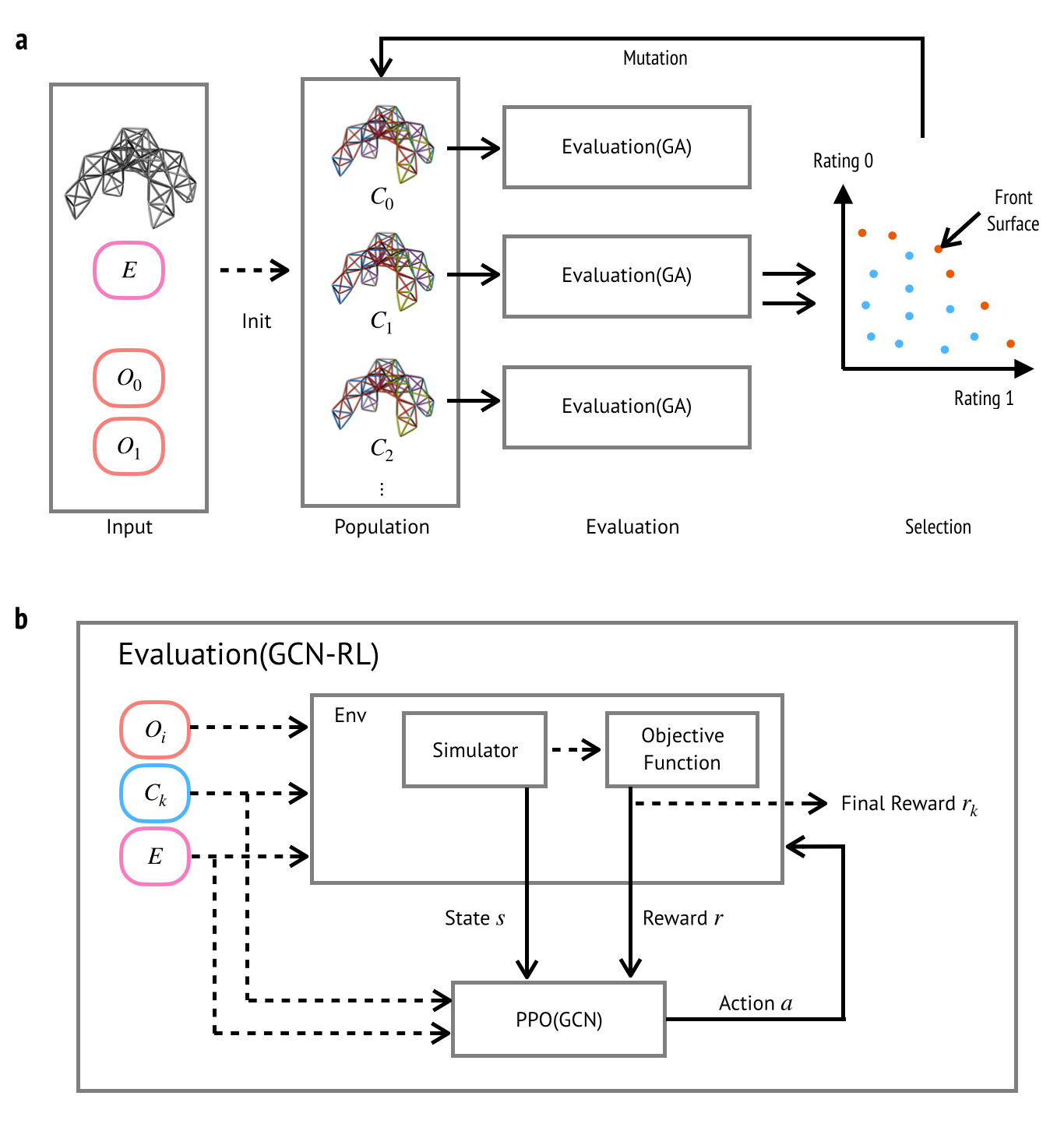}}
\caption{Computational pipeline of channel grouped VGT co-design.}
\label{fig3}
\end{center}
\end{figure}

\subsection{Overview}
We introduce a co-design algorithm which uses GA to optimize $C$ and RL to optimize $\Pi$ (Figure \ref{fig3}). 
GA is first used to initialize a generation $G$ of channel assignments $C_0$,$C_1$, ...$C_{n_g - 1}$, where $n_g$ is the population size of one generation. Each channel assignment is bundled with a control sequence $\xi$ that includes $n_s \times n_\gamma$ Boolean values, each indicating the on/off state of the $i_\gamma \text{th}$ channel at the $i_s$ time step. 

In each generation, every pair of {$C_k, \xi_k$} are tested by a simulator which outputs a rating vector $R_k$. At the end of the generation, a selection function (Appedix \ref{nsga}) following NSGA-II[\cite{nsga2}] keeps the population $\bar{G}$ with the best performance and discards the rest. The missing population will be filled by duplicating and mutating $\bar{G}$ through a customized mutation function (Appendix \ref{init}) that assures the channel connectivity and the symmetry of the truss. We also introduce an elite pool mechanism (Appendix \ref{elite}) that temporarily keeps elite populations aside to avoid exploitation caused by the domination of elites. 

However, $\xi$ is a fixed sequence that only applied to the truss with a specific initial condition and at a specific environment. Once the initial condition changes, $\xi$ needs to be retrained. Furthermore, errors will accumulate in a long trajectory. Instead, we train a RL model as a closed-loop control policy $\pi_k$ on top an elite $C_k$ in the last GA generation.

% The channel assignment $C$ are discrete both in their values and the space. We use GA as the non-differentiable method to optimize $C$. 
%     Trusses have a hard requirement on the channel connectivity, where a random change in $C$ might break the connectivity and render the design physically impractical. 
    
%     Considering the aesthetics, we limit the truss to be mirrored to a vertical mirror plane. Meanwhile, the channels are required to have the symmetry as well to ensure a symmetric motion pattern. 
    
%     We reduce the graph of the truss to a semi-graph and obtain $\hat{T}, = {\hat{E}, n_\gamma}$, limiting the search space of the GA within the half of the truss (Figure \ref{fig4}).
%     We use NSGA-II as the selection function for the performance on multiple objectives. 

% RI has been proven to be effective on control optimization of a wide range of robotic systems.

\subsection{Channel Symmetry Requirement}
If the input truss is mirrored, a channel symmetry requirement is enforced(Figure \ref{fig4}).With this requirement, each channel is either mirrored to itself or mirrored to another channel. For example, each pair of mirrored edges are either in the same channel or in two mirrored channels. By default, half of the channels will be self-mirrored and the other half are inter-mirrored. In such a way, we assure the symmetry of the channels and implicitly assure the possibility of symmetric motions.

A mirroring function maps a half graph $\hat{A}$ to the whole graph $\hat{A}$. The \textit{initialization} and \textit{mutation} steps are performed in the half graph and mirrored to the whole graph when simulating. Other parts remains the same.

\subsection{Genetic Algorithm}
A GA training process consists of two nested loops. The outer loop is for exploration and the inner loop for exploitation. An evolving pool is updated at every exploitation step, and an elite pool is updated every exploration step. ($C, \xi$) pairs are initialized in the evolving pool and improved by iteratively applying selection and mutation (Appendix \ref{init}) functions. After certain generations(20 in this work), survived pairs will be put into an elite pool and a new evolving pool will be initialized. When the elite pool is full, elite pairs will be put back into the evolving pool to further evolve. See details in (Appendix \ref{ga}).

\subsection{Reinforcement Learning}

We use PPO \cite{ppo} as the reinforcement learning method to train the control policy and use PytorchRL \cite{pytorchrl} for implementation. We use two multi-layer perceptrons(MLP) each with two linear layers for the actor and critic network.

\subsubsection{Observation}
The observation is calculated at every simulation time step to inform the controller the state of the truss. As mentioned above, a state vector is $s_t = \{V_t, U_t, {V_t}', {U_t}', k_t\}$. Kinematics information in the model space is location invariant and gives controller less noise than the information in the world space \cite{gcnphysics}. To calculate the relative node positions and velocities, for every truss, we define a beam in the middle as the center beam. The center beam determines the front orientation $q_c$ as well as the center location $p_c$ of the truss, which we use to calculate the relative positions ${V_t}'$ and velocities ${U_t}'$ by translating and rotating $V_t$, $U_t$ based on $q_c$, $p_c$.

\subsubsection{Action}
At each time step, an action vector is generated from the controller and sent to the simulator. For a truss with $n_\gamma$ channels, action is a Boolean vector with $n_\gamma$ digits. The action vector is encoded as a binary value to a decimal value within $[0, 2^{n_\gamma})$. Accordingly, The output linear layer has $n_\gamma$ outputs. When using deterministic policy, PPO takes the largest output as the decimal and converts it to the action vector.

\subsubsection{Reward}
Rewards are calculated by customized reward function. For each task, once the simulation reaches a maximum time steps or a target objective is achieved, the simulation finishes and gives a task specific reward. Before finishing, simulator gives 0 reward. For example, the distance of locomotion and height change are rewards for moving forward and lowering bodies. For turning left and tilting the top, a negative angle difference will be calculated based on the target and final orientations.

% The truss has an underlying graph structure connecting all the joints, with truss beams defined as the edges, and joints of the trusses defined as nodes. To utilize the hidden information within the graph, graph neural network \cite{gcnsemi} is adopted for the actor and critic network in PPO. Specifically, each node in GCN initially stores the position and velocity information of each corresponding joint in the truss.

% A two-layer GCN is used for the actor network. We use another GCN with the same structure for the critic network followed by a MLP layer that takes in the output of GCN and generates the estimated $V$ value. An action is sampled from a categorical distribution parameterized by the output vector of the actor GCN.

% We use PytorchRL \cite{pytorchrl} for implementing PPO and PyG \cite{gcn_github} for implementing GCN.

\section{Preliminary Result}
We demonstrate the usability and the performance of our method with a robotic table.  We envision a scenario where every piece of furniture and daily objects can be robotic and have extra morphing functionalities in addition to their conventional static state. Here, a robotic table is trained to walk towards people who have limited mobility, lower itself for kids, or tilt the top in an angle for painters. in order to achieve the target design features, four objective functions were defined during the training process: moving forward, turning left by 90°, lowering the height, and tilting the tabletop. We design the truss structure of the table and fix the relative position of the top four nodes to mimic a fixed tabletop in the simulator. 

\begin{figure}[H]
% \vskip 0.2in
\begin{center}
\centerline{\includegraphics[width=\columnwidth]{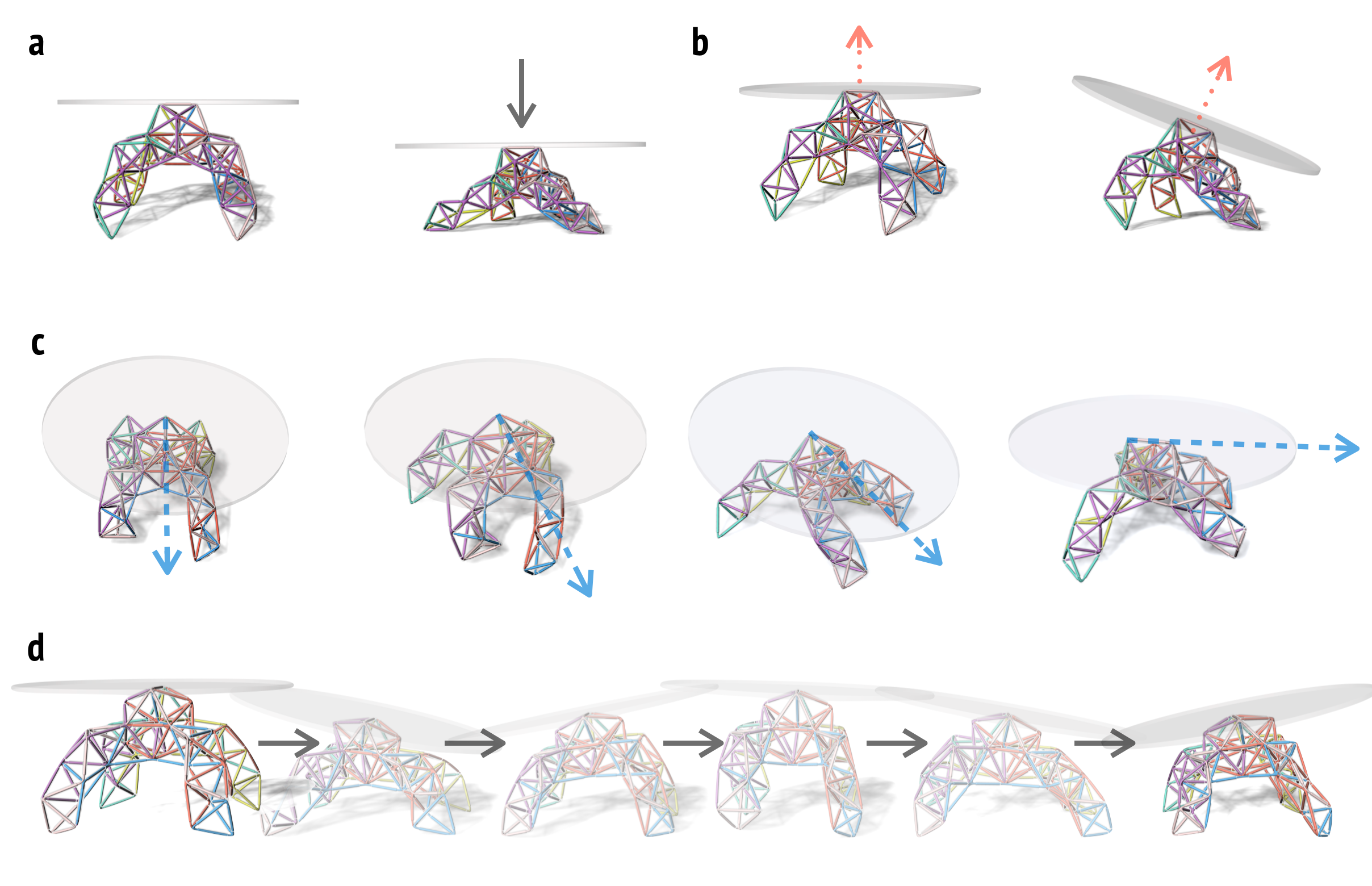}}
\caption{Four motions of a robotic table, (a) lowering the height, (b) tilting the table top, (c) rotating, and (d) locomoting forward.}
\label{fig5}
\end{center}
% \vskip -0.2in
\end{figure}

We first use GA to optimize ($C, \xi$) for 1000 generations. We then pick a $C$ from the last generation and train RL for 600 model updates. The experiment is run on a server with 64 cores for 2.4 hours. Figure \ref{fig6} shows that GA effectively optimizes the design towards four objectives. Although RL improves the performance only by a small value, control policy by RL is adaptable to a wider range of initial condition and is able to adjust the control on the fly to minimize the accumulated error. Further research needs to be conducted to test the generalizability of RL under different initial conditions, terrain conditions and external disturbance. We visualizes the design and the trajectory of the robotic table for four tasks using the training result by RL. (Figure \ref{fig5}).

% \section{Implementation}

% Simulator
% A mass-spring based simulator[] is used, implemented with pyTorch. Extra rigid springs with high stiffness are added to points that have fixed relative distance. For example, the attachment points for the tabletop have six extra rigid springs. 

% We trained on a 64-core V100 GPU.

\section{Discussion}

The preliminary result shows that our method is able to generate useful channel designs and controls for a complex truss with multiple diverse objectives. An immediate next step is to test the generalization ability of the method. For example, testing the method with trusses of various shapes, diverse objective functions, varying terrain conditions and external disturbance. To improve the generalization ability, we can apply graph convolutional neural network \cite{gcnsemi} to utilize the underlying graph topology information. Currently, the initial truss topology is pre-designed, but it might not be the optimal topology for the given tasks. To further improve the truss topology, a generative design such as a graph heuristic search \cite{zhao2020robogrammar} might optimize the truss topology together with the design and control.

\bibliography{manuscript}
\bibliographystyle{icml2022}

%%%%%%%%%%%%%%%%%%%%%%%%%%%%%%%%%%%%%%%%%%%%%%%%%%%%%%%%%%%%%%%%%%%%%%%%%%%%%%%
%%%%%%%%%%%%%%%%%%%%%%%%%%%%%%%%%%%%%%%%%%%%%%%%%%%%%%%%%%%%%%%%%%%%%%%%%%%%%%%
% APPENDIX
%%%%%%%%%%%%%%%%%%%%%%%%%%%%%%%%%%%%%%%%%%%%%%%%%%%%%%%%%%%%%%%%%%%%%%%%%%%%%%%
%%%%%%%%%%%%%%%%%%%%%%%%%%%%%%%%%%%%%%%%%%%%%%%%%%%%%%%%%%%%%%%%%%%%%%%%%%%%%%%
\newpage
\appendix
\onecolumn

\section{Genetic Algorithm Pipeline}
\label{ga}

    In the first exploitation step, the evolving pool of $n_{g}$ designs $(C, \xi)$ are generated through an \textit{initialization} function, which either randomly creates $(C, \xi)$ pairs or duplicate from the elite pool if it is full. Ratings are generated through the \textit{evaluation} function which simulates the truss with ${C, \xi}$. Based on the ratings, a \textit{selection} (Appendix \ref{nsga}) function sorts and keeps a subset of the population to maximize the general population performance as well as the diversity. 
    
    In every other exploitation step, the preserved population will be duplicated and mutated to fill the evolving pool through a \textit{mutation} (Appendix \ref{init}) function before the \textit{evaluation}.
    
    At the end of an exploitation loop, the preserved population will be added to an elite pool which has the same maximum size as the evolving pool. When the elite pool is full, elite population will be duplicated for initializing the evolving pool and the elite pool is cleared. After an exploitation step, the better elites will be selected and put back to the elite pool.
    
    In such a way, the existing elite pool is continuously improved while the new population will have the chance to improve in each exploitation loop before being defeated by the elites.

\section{Initialization and Mutation with Channel Connection Constraint}
\label{init}

Different from traditional GA where initialization and mutation are giving random value to randomly chosen digits, the digits in this problem are the channel assignment on beams, which is constrained by the channel connection rule. Beams of the same channel must be connected through joints and a random change might separate a channel and renders the solution impractical. Therefore, we use a customized channel-growing algorithm to initialize and mutate the solution.

At the initialization stage (Appendix \ref{alg:intialization}), the algorithm assigns $n_c$ edges with different channels. To assure the connectivity, a self-mirrored channel will be initialized with one self-mirrored edge. The algorithm keeps track of all the unassigned edges that are connecting to the assigned edges as $E_{uc}$. It randomly chooses one edge from $E_{uc}$ and randomly assigns a channel $\gamma$ that it is incident to the edge. The previous step is repeated until all the edges are assigned.

The \textit{mutation} function (Appendix \ref{alg:mutation}) is similar to \textit{initialization}. It keeps track of all the edges that are connected with more than one channel as $E_c$. Every time, it selects one edge $e_c$ from $E_c$ and picks one different incident channel $\bar{c}$ and reassigns $\bar{c}$ to $e_c$. It checks the connectivity of all the subgraphs with breadth-first search and reverts the reassignment if the connectivity is broken. It repeats the previous step until finding an assignment that does not break the connectivity.

\section{Multi-Objective Selection Function}
\label{nsga}
We adopted NSGA-II \cite{nsga2} as the selection function to keep the multi-objective performance. 

This method first preserves solutions at the front hypersurface where no solution is worse than another solution on all objectives. Within the front hypersurface, a crowdedness distance(CD) is calculated based on how crowded the solutions are in the local performance space. If the total number of preserved solutions exceeds the threshold, the solutions with lower CD values will be preserved. Different from multi-objective functions with weighted sum that might lead to solutions that are not good at any objectives, this method improves the ratings of each objective while keeping the diversity of the genes. Meanwhile, It does not require a weight assignment to every objective.

\section{Elite Pool Mechanism}
\label{elite}

Every fixed number of generations, the population remaining on the front hypersurface will be added to an elite pool, and the population will be cleared and reinitialized. When the elite pool reaches the size of the population, the population in the elite pool will be put back to the population. This mechanism allows GA to explore more types of solutions by re-initialization but at the same time exploit the best solutions by repetitively mutating the preserved population.

\begin{figure}[ht]
% \vskip 0.2in
\begin{center}
\centerline{\includegraphics[width=\columnwidth]{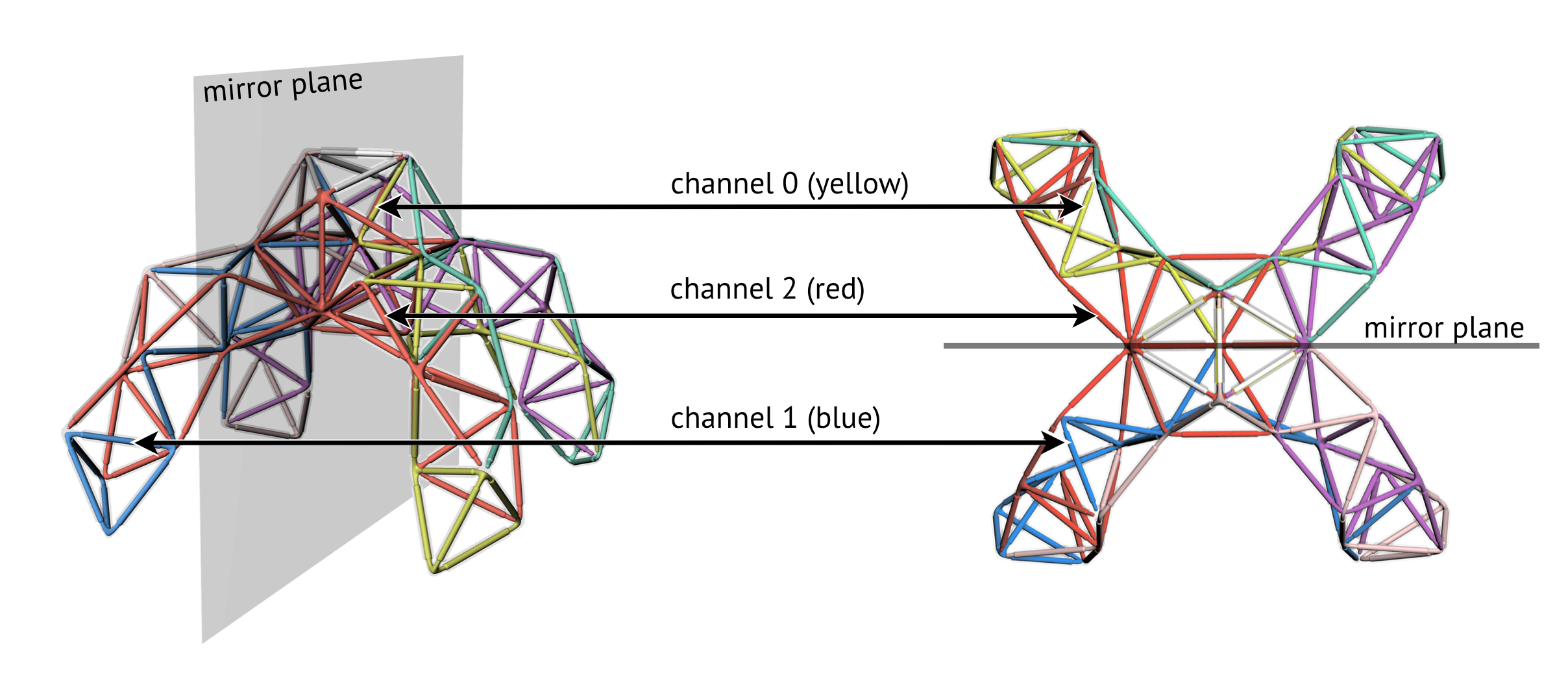}}
\caption{Channel symmetry constraint. Left: Perspective view of the truss of a robotic table. Right: Top view of the robotic table truss. Channel 0 and channel 1 are symmetric to each other with regard to the mirror plane. Channel 2 is symmetric by itself.}
\label{fig4}
\end{center}
% \vskip -0.2in
\end{figure}

\begin{algorithm}[tb]
  \caption{Initialization}
  \label{alg:intialization}
 
\begin{algorithmic}
    \STATE $n_\gamma$: number of channels
    \STATE Total channels $\Gamma=\{\gamma_0, \gamma_1, ... \gamma_{n_\gamma - 1}\}$, where $\gamma_i = i$ for simplicity.
    \STATE Total channels of a halfgraph $\hat{\Gamma} = \{ \hat{\gamma_0}, \hat{\gamma_1}, ... \gamma_{\hat{n_\gamma} - 1} \}$
    \STATE $\phi: \gamma \mapsto \gamma_{m}$, where $\gamma_m$ is the channel mirrored to $\gamma$.
    \STATE $E = \{e_0, e_1, ... e_{n_e - 1}\}$, where $e_i = {v_{i0}, v_{i1}}$
    \STATE $\hat{E} = \{ \hat{e_0}, \hat{e_1}, ... \hat{e_{n_e - 1}} \}$, edges in the half graph.
    \STATE $\psi: e \mapsto e_m$, where $e_m$ is the edge mirrored to $e$.
    \STATE $\hat{E_u}$: edges on the half graph with no channel assignment.
    \STATE $E_m$: edges that are self-mirrored.
    \STATE $C={c_0, c_1, ... c_{n_e - 1}}$, $c_i \in \Gamma$, the channel assignment of each edge, initialized with -1.
    
    \FOR{ $\gamma = 0$ {\bfseries to} $n_e - 1$}
        \STATE $\gamma_m \leftarrow \phi(\gamma)$
        
        \IF{ $\gamma_m == \gamma$}
            \STATE select $e$ from $\hat{E_u} \cap E_m$
            \STATE $c_e \leftarrow e$ 
        \ELSE
            \STATE select $e$ from $\hat{E_u} \cap (\hat{E} - E_m)$
            \STATE $e_m \leftarrow \psi(e) $
            \STATE $c_e \leftarrow \gamma $
            \STATE $c_{e_m} \leftarrow \gamma_m$
        \STATE remove $e$ from $\hat{E_u}$
        
        \ENDIF
    \ENDFOR

    \WHILE{$\hat{E_u}$ is not empty}
        \STATE $\hat{E_{adj}} \leftarrow$ \text{edgesConnectingChannels}($C$, $\hat{E}$)    %\COMMENT{find edges that are connected to edges assigned with channels}
        \STATE select $e$ from $\hat{E_{adj}}$
        \STATE $\Gamma_{inc} \leftarrow$ \text{channelsIncidentEdge}($C$, $e$)      %\COMMENT{find all channels connected to the edge}
        \STATE select $\gamma$ from $\Gamma_{inc}$
        \STATE $c_e \leftarrow \gamma$
        \STATE remove $e$ from $\hat{E_u}$
    \ENDWHILE

\end{algorithmic}

\end{algorithm}

\begin{algorithm}[tb]
  \caption{Mutation}
  \label{alg:mutation}
 
\begin{algorithmic}
    \STATE $\hat{E_{c}} \leftarrow$ \text{edgesConnectingMultipleChannels}($C$, $\hat{E}$)      %\COMMENT{find edges connected to more than one channels}
    \WHILE{True}
        \STATE select $e$ from $\hat{E_{c}}$
        \STATE $\Gamma_{inc}\leftarrow$ \text{channelsIncidentEdge}($C$, $e$)      %\COMMENT{find all channels connected to the edge}
        \FOR{$\gamma$ in $\Gamma_{inc}$}
            \IF{\text{channelsConnected}($C$, $\hat{E}$, $e$, $gamma$)}     %\COMMENT{check if the channels are connected after reassigning $e$ with $gamma$}
            
                \STATE $\gamma_{e} \leftarrow \gamma$       %\COMMENT{reassign a channel to the edge}
                \STATE BREAK
            \ENDIF
        \ENDFOR
    
    \ENDWHILE

\end{algorithmic}

\end{algorithm}

\begin{figure}[H]
% \vskip 0.2in
\begin{center}
\centerline{\includegraphics[width=\columnwidth]{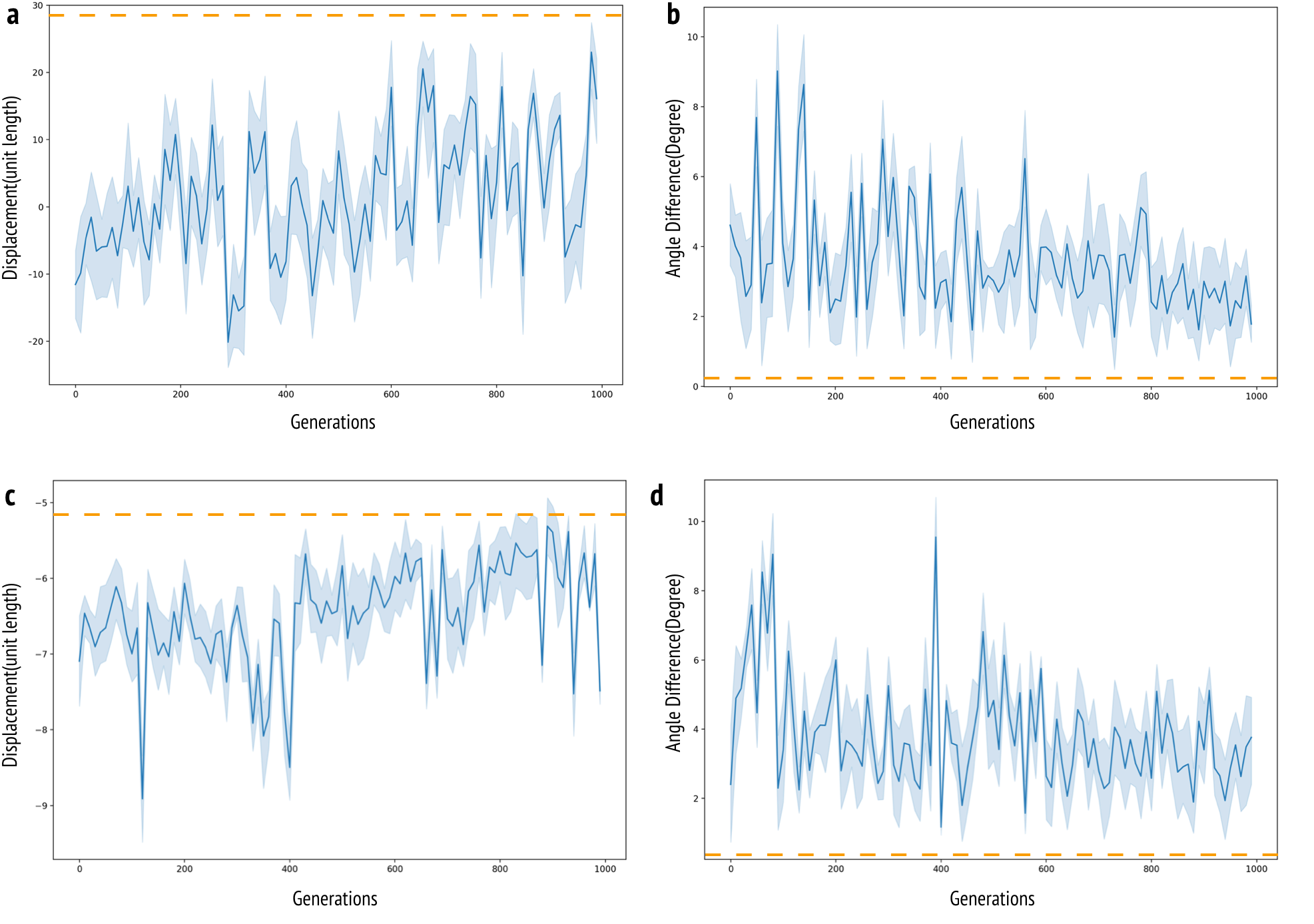}}
\caption{Blue curves are the performance of GA algorithms where the repetitive switching between evolving pool and exploitation pool causes the performance to increase in fluctuation. Orange dashed lines indicate the eventual performance by reinforcement learning trained on top of a channel design from the last generation of GA. Four tasks are tested, including (a) moving forward, (b) rotating by 90 degrees. (c) lowering the height, (d) tilting the table top.}
\label{fig6}
\end{center}
% \vskip -0.2in
\end{figure}

% You can have as much text here as you want. The main body must be at most $8$ pages long.
% For the final version, one more page can be added.
% If you want, you can use an appendix like this one, even using the one-column format.

%%%%%%%%%%%%%%%%%%%%%%%%%%%%%%%%%%%%%%%%%%%%%%%%%%%%%%%%%%%%%%%%%%%%%%%%%%%%%%%
%%%%%%%%%%%%%%%%%%%%%%%%%%%%%%%%%%%%%%%%%%%%%%%%%%%%%%%%%%%%%%%%%%%%%%%%%%%%%%%

\end{document}